\pdfoutput=1

\documentclass[11pt]{article}

\usepackage {EMNLP2023}

\usepackage{times}
\usepackage{latexsym}
\usepackage{amsmath}
\usepackage{graphicx}
\usepackage{comment}
\usepackage[T1]{fontenc}

\usepackage[utf8]{inputenc}

\usepackage{microtype}

\usepackage{inconsolata}

\title{Re-Ranking Step by Step:

Investigating Pre-Filtering for Re-Ranking with Large Language Models}

\author{Baharan Nouriinanloo \and Maxime Lamothe \\
        Polytechnique Montréal \\
        Quebec, Canada \\
        \{baharan.nouriinanloo, maxime.lamothe\}@polymtl.ca}

\begin{document}
\maketitle
\begin{abstract}
Large Language Models (LLMs) have been revolutionizing a myriad of natural language processing tasks with their diverse zero-shot capabilities. Indeed, existing work has shown that LLMs can be used to great effect for many tasks, such as information retrieval (IR), and passage ranking. However, current state-of-the-art results heavily lean on the capabilities of the LLM being used. Currently, proprietary, and very large LLMs such as GPT-4 are the highest performing passage re-rankers. Hence, users without the resources to leverage top of the line LLMs, or ones that are closed source, are at a disadvantage. In this paper, we investigate the use of a pre-filtering step before passage re-ranking in IR. Our experiments show that by using a small number of human generated relevance scores, coupled with LLM relevance scoring, it is effectively possible to filter out irrelevant passages before re-ranking. Our experiments also show that this pre-filtering then allows the LLM to perform significantly better at the re-ranking task. Indeed, our results show that smaller models such as Mixtral can become competitive with much larger proprietary models (e.g., ChatGPT and GPT-4).
\end{abstract}

\section{Introduction}

Large Language Models (LLMs) such as GPT-4~\cite{achiam2023gpt} and Mixtral~\cite{jiang2024mixtral} have been revolutionizing natural language processing tasks with their diverse zero-shot capabilities. Through extensive pretraining on various large-scale textual sources—such as web pages, research articles, books, examples, and code—these systems have shown remarkable natural language capabilities. As a result, their responses are increasingly human-like and closely aligned with human intentions~\cite{zhu2023large}. 
Recently, as LLMs have become more capable, extensive research has been conducted on their use and effectiveness in information retrieval (IR) systems~\cite{zhu2023large}. 

A typical information retrieval system comprises multiple components organized into a processing pipeline. This pipeline features two primary stages: the retriever and the reranker~\cite{lin2022pretrained}. While the retriever selects the most relevant passages from a large-scale corpus, the re-ranker focuses on re-ordering (i.e., re-ranking) the candidate passages, using their relevance. 
Each component can thus be optimized for its given task.

The advent of large language models (LLMs) has impacted the information retrieval (IR) pipeline in many ways. While research in this field has mainly focused on the use of LLMs in the first stage of the pipeline~\cite{zhu2023large} (i.e., the retriever), the investigation of LLMs for the re-ranking stage remains a relatively new challenge. Recently, some research has focused on using LLMs for zero-shot re-ranking, leading to significant improvements~\cite{sun2023chatgpt}. 

Despite recent advances in retrieval methods, some retrieved passages are likely to be unrelated to the query. Generally, all of the retrieved passages from the retriever (typically BM25~\cite{lin2021pyserini}) are passed to the LLM-based re-ranker, to generate the ranked list of passages based on their relevance to the query. Passing irrelevant and distracting passages to the LLM-based re-ranker can mislead it, leading it away from relevant passages and negatively impacting the ranking process~\cite{cuconasu2024power, yoran2023making}.

To investigate the impact of these irrelevant passages and mitigate their effects, we propose a novel LLM-based pre-filtering method that filters out irrelevant passages before they are given to the re-ranker. We design a prompting strategy for our method which instructs the open-source LLM to generate a relevance score for each passage based on its relevance to the given query in the range of 0 to 1. 
Then, using a sample of generated scores, we establish a specific threshold for passage filtering. Using this threshold we can then pre-filter any new passage. Passages that exceed this threshold are retained as relevant passages and forwarded to the re-ranker, while those falling below the threshold are discarded.
By implementing this straightforward process, only relevant passages are passed to the ranker, significantly reducing the overall number of passages in the ranker. To investigate the usefulness of our proposed approach we focus on the following two research questions:
\vspace{-0.5em}
\begin{itemize}
    \item[] \textbf{RQ1:} Can existing expert knowledge be used to help LLMs filter out irrelevant passages?
    \item[] \textbf{RQ2:} Does filtering out irrelevant passages before re-ranking improve the results of an LLM re-ranker?
\end{itemize}
\vspace{-0.5em}

To answer \textbf{RQ1} we investigate prompting Mixtral-8x7B-Instruct (with 4-bit quantization) to assign a quantitative value to the relevance of passages retrieved by BM25. We then leverage existing expert knowledge (i.e., qrels in the TREC and BEIR datasets), to determine a relevance value below which passages should be deemed irrelevant (i.e., a relevance threshold). We evaluate our approach on three datasets (TREC-DL2019, TREC-DL2020, and four BEIR tasks) and find that it is generally possible to find a threshold value--using F1 score--that maximizes the relevance of the retrieved passages. We also find that this threshold value appears mostly stable (around 0.3, or its inverse 0.7) across all of the tested datasets.

To answer \textbf{RQ2}, we investigate the use of a pre-filtering step--to filter out irrelevant passages--before re-ranking passages with Mixtral-8x7B-Instruct (loaded with 4-bit quantization). Again, we evaluate our approach on three datasets (TREC-DL2019, TREC-DL2020, and four BEIR tasks) and find that the use of a pre-filtering step significantly improves the resulting re-ranking of passages. Indeed, after using our pre-filtering step, a limited model such as Mixtral-8x7B-Instruct (loaded with 4-bit quantization) can become competitive with-- and in one case surpass--much larger, and resource intensive, models such as GPT-4.
Our paper presents three distinct contributions:
\begin{itemize}
    \item We show that it is possible to use expert knowledge as feedback to LLMs to identify irrelevant passages in IR.
    \item We present a novel approach to improve LLM-based passage re-ranking.
    \item We show that while our approach is model agnostic, by using it, resource-constrained LLMs can become competitive with resource-intensive LLMs that do not use our approach.
\end{itemize}

\section{Related Work}

Recently, large language models (LLMs) have significantly impacted various research fields, including information retrieval.~\cite{zhu2023large}. Several approaches have been put forward to use LLMs as re-rankers in the information retrieval pipeline. These LLM-based re-rankers can be categorized into supervised and unsupervised methods. 

Existing supervised re-rankers can be categorized as: (1) encoder-only, (2) encoder-decoder, and (3) decoder-only. monoBERT~\cite{nogueira2019multi} is an encoder-only re-ranker based on the BERT-large model. It leverages BERT’s contextualized embeddings to enhance document re-ranking performance and is optimized using cross-entropy loss. monoT5~\cite{nogueira-etal-2020-document} is an encoder-decoder-based re-ranking model designed for information retrieval tasks. It leverages the T5 (Text-To-Text Transfer Transformer) architecture to generate relevance scores for query-document pairs by treating the ranking task as a sequence-to-sequence generation problem. RankT5~\cite{zhuang2023rankt5} is another encoder-decoder-based re-ranking model which calculates the relevance score for a query-document pair and optimized the ranking performance with pairwise or listwise ranking losses. RankLLaMA~\cite{ma2023fine} is a decoder-only re-ranker model that focuses solely on the output generation phase. It uses the last token representation for relevance calculation. Compared to prior work, we use Mixtral, a recent LLM, as a re-ranker coupled with a novel step--pre-filtering irrelevant passages--to improve the re-ranking process.

As the size of LLMs scales up, it becomes difficult to fine-tune the re-ranking model. Recent efforts aim to tackle this challenge by prompting LLMs to directly enhance document re-ranking in an unsupervised manner. Generally, there are three main methods for using LLMs in zero-shot re-ranking tasks: Pointwise ~\cite{sachan2022improving,liang2022holistic}, Listwise~\cite{sun2023chatgpt, ma2023zero}, and Pairwise~\cite{qin2023large}. In this paper, we use a listwise approach as it strikes a balance between efficiency and effectiveness.

There are two popular methods for prompting LLMs to rank documents in a pointwise manner: relevance generation~\cite{liang2022holistic} and query generation~\cite{sachan2022improving}. The Unsupervised Passage Re-ranker (UPR)~\cite{sachan2022improving} is a pointwise approach based on query generation. In this approach, LLMs are prompted to produce a relevant query for each candidate document. The documents are then re-ranked based on the likelihood of generating the actual query.
Relevance Generation (RG)~\cite{liang2022holistic} is another pointwise approach based on relevance generation. In this method, LLMs are prompted to generate whether the provided candidate document is relevant to the query, with this process repeated for each candidate document. Subsequently, these candidate documents are re-ranked based on the normalized likelihood of generating a "yes" response. In this paper we add a new step (i.e., pre-filtering) to the information retrieval pipeline, before the re-ranker. We then leverage existing work to re-rank our pre-filtered passages.

The main goal of the listwise approach ~\cite{sun2023chatgpt, ma2023zero} is to directly rank a list of candidate documents for a given query and generate a ranked list of document labels based on their relevance to the query. In this method, the query and a list of documents are directly inserted into a prompt. Due to the prompt length constraints of LLMs, this approach employs a sliding window method, which involves re-ranking a window of candidate documents. This process starts from the bottom of the original ranking list and progresses upwards. It can be repeated multiple times to achieve an improved final ranking, allowing for early stopping mechanisms to target only the top-K rankings, thereby conserving computational resources.

In pairwise methods ~\cite{qin2023large}, LLMs are given a prompt consisting of a query and a document pair. They are then instructed to identify the document with higher relevance. To rerank all candidate documents, aggregation methods like AllPairs are used. AllPairs first generates all possible document pairs and then aggregates a final relevance score for each document. To expedite the ranking process, efficient sorting algorithms, such as heap sort and bubble sort, are typically employed. These algorithms use efficient data structures to selectively compare document pairs and elevate the most relevant documents to the top of the ranking list, which is particularly useful in top-k ranking. Although effective, pairwise methods suffer from high time complexity. To address this efficiency problem, a setwise approach~\cite{zhuang2023setwise} has been proposed, which compares a set of documents simultaneously and selects the most relevant one. This approach enables sorting algorithms, such as heap sort, to compare more than two documents at each step, thereby reducing the total number of comparisons and speeding up the sorting process.

\section{LLM-based Pre-Filtering}
\label{sec:Methodology}
In this section, we propose a new LLM-based pre-filtering step to score passages based on their relevance to the query and filter out irrelevant passages before any re-ranking is conducted. Below, we explain the significance of this step, describe the prompting strategy used for generating relevance scores, outline the process of analyzing the generated scores, and explain how we set the thresholds.

\begin{figure*}[t]
    \centering
     \includegraphics[width=.8\linewidth]{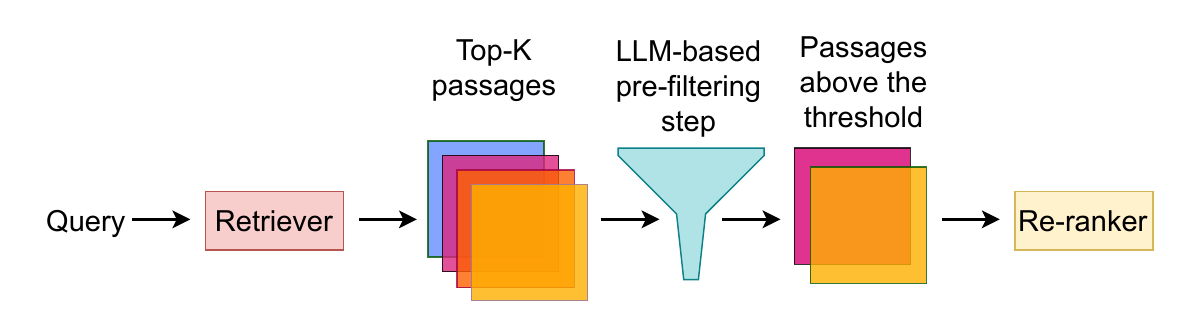}
    \caption{ The role of the pre-filtering step in the information retrieval pipeline. }
    \label{ranking_pipeline}
\end{figure*}

Our proposed pre-filtering step is straightforward yet efficient. The main goal of this step is to filter irrelevant passages before passing them to a re-ranker, thus decreasing the total number of passages that require re-ranking. After the initial retrieval stage (e.g., based on BM25~\cite{lin2021pyserini}) retrieves a set of passages for a given query, each passage is evaluated by an LLM-based filter to determine its relevance to the query. This filter, using the language understanding capabilities of LLMs, assigns a relevance score to each passage (e.g., from 0 to 1 where 0 denotes a completely irrelevant passage, and 1 denotes a fully relevant one). 
Then, a threshold is set based on these scores. Passages with scores at or above the threshold are passed to the re-ranker, while those with scores below the threshold are discarded. Using the pre-filtering step, the number of noisy passages that can misguide the re-ranker decreases, leading to improved performance for the re-ranker. Figure \ref{ranking_pipeline} illustrates the role of the pre-filtering step in the information retrieval pipeline.

\vspace{-0.5em}

\subsection{Prompt Design for Score Generation}
To design our prompt, we use two well-known prompting methods:
\vspace{-0.5em}
\begin{itemize}
   \item[] \textbf{Chain-of-Thought (CoT)}~\cite{wei2022chain, kojima2022large}: This method allows LLMs to produce intermediate reasoning steps explicitly before generating the final answer.
   \vspace{-0.5em}
   \item[] \textbf{Plan-and-Solve (PS)}~\cite{wang2023plan}: This method consists of designing a plan to divide a task into smaller subtasks and then carrying out the subtasks according to the plan.
\end{itemize}
Our proposed zero-shot prompt is a combination of both of these reasoning methods:
\par
\textit{“Grasp and understand both the query and the passages before score generation. Then, based on your understanding and analysis quantify the relevance between the passage and the query. Give the rationale before answering.”}

In the first part of the prompt, we devise a plan for the LLM to understand both the query and the passage, and then, based on its analysis, generate a relevance score. In the second part, we include a sentence asking the LLM to explain the rationale behind the score generation. By designing this plan and instructing the LLM to provide a rationale before responding, its understanding of the query and the passage is incrementally enhanced. Our manual analysis of multiple different prompts led us to believe that this step-by-step approach results in the generation of more accurate relevance scores.

\subsection{Analyzing Relevance Scores}

To give context to the relevancy scores generated by LLMs for the passages, they should be compared with an existing baseline. In our experiments, we make use of the relevancy levels in the query relevance judgments (qrels) file for each passage. These relevancy levels have somewhat similar interpretations across different datasets; however, details can be different. In all cases, higher scores indicate greater relevance between the query and the passage. However, the interpretation of a \textbf{level [{}1{}]} score differs: in some datasets, a score of \textbf{[{}1{}]} is considered relevant; in others, irrelevant; and in some others, it is interpreted as partially relevant. 
We categorize the datasets based on the the interpretation of a \textbf{level [{}1{}]} score as follows:

\begin{itemize}
\item Datasets where a relevancy level of \textbf{[{}1{}]} is interpreted as either relevant (e.g., Touche~\cite{thakur2021beir}) or partially relevant (e.g., Covid~\cite{thakur2021beir}), a relevancy level of \textbf{[{}1{}]} or higher is considered relevant.
\item Datasets where a relevancy level of \textbf{[{}1{}]} is interpreted as not relevant (e.g., TREC-DL19, TREC-DL20~\cite{craswell2020overview,craswell2021overview2020}), a relevancy level of \textbf{[{}1{}]} or lower is considered irrelevant. In these datasets the nDCG scores use a gain value of 1 for related passages. 

\end{itemize}

\subsection{Setting a Relevance Threshold}
\label{sec:threshold}
Since the scores generated by LLMs are decimal numbers between 0 and 1, we convert these scores to be comparable with the integers that represent relevancy levels. Thus, we set a threshold to replace the decimal scores with the following values:

\begin{equation}
S_{\text{pre}}\ = 
\begin{cases} 
1 , & \text{if the score at the threshold} \\ 
1 , & \text{if the score above the threshold} \\ 
0, & \text{if the score below the threshold} 
\end{cases}
\end{equation}

Where passages with a relevancy score at or above this threshold are considered relevant and assigned a score of 1, while those below the threshold are given a score of 0. Next, the four elements of the confusion matrix—true negatives, true positives, false positives, and false negatives—are calculated. This is done by comparing the \(S_{\text{pre}}\) for each passage with the relevancy levels in the qrels file. Since not all passages in each dataset have relevancy levels assigned, we only use the passages with relevancy levels in the qrels file to compute these four elements. After the threshold value is selected, even passages without relevancy levels can then be considered, as the LLM can still generate scores for them, and their scores can thus be replaced based on the previously defined threshold. Based on these values, Precision, Recall, and F1 Score are computed for each threshold and the threshold with the highest F1 Score is selected.
In the context of IR systems, these elements are defined as follows:
\vspace{-0.5em}
\begin{itemize}
    \item[] \textbf{True Negative (TN):} Both the qrels files and the LLM classify the passage as irrelevant.
    \item[] \textbf{True Positive (TP):} Both the qrels files and the LLM classify the passage as relevant.
    \item[] \textbf{False Positive (FP):} The qrels files classify the passage as irrelevant, but the LLM classifies it as relevant.
    \item[] \textbf{False Negative (FN):} The qrels files classify the passage as relevant, but the LLM classifies it as irrelevant.
\end{itemize}
\vspace{-0.5em}

While the initial threshold value is randomly selected within the predefined range of 0 to 1, our analysis suggests that this selection should align with the interpretation of relevancy levels in the qrels files. For datasets where a relevancy level of \textbf{[{}1{}]} is interpreted as either relevant or partially relevant, a smaller threshold value should be chosen compared to datasets where a relevancy level of \textbf{[{}1{}]} is interpreted as not relevant. In the first case, even passages with a low relevance, represented by a relevancy score of \textbf{[1]}, are considered relevant. In this case, the scores generated by the LLMs for those passages are generally lower than those for passages with a higher level of relevancy. Therefore, the selected threshold should be more inclusive. Conversely, for the second case where a relevancy level of \textbf{[1]} is interpreted as not relevant, the threshold should be higher to include only the more relevant passages.
Additionally, this suggests that when testing thresholds other than the initial one, comparing it with a slightly higher and a slightly lower value and computing the F1 score for these thresholds is sufficient to identify the trend of the other values and select the best one.






\subsection{The Advantages of Pre-Filtering}

As the primary objective of IR systems is to accurately detect information that fully or partially matches the user's query, identifying relevant and irrelevant passages is crucial. Therefore, the main goal of our proposed step is to precisely detect both relevant and irrelevant passages. By selecting the best F1 score we aim to optimize the threshold to increase the number of True Positives (TPs) and True Negatives (TNs) while decreasing the number of False Positives (FPs) and False Negatives (FNs). By removing distracting passages, the total number of initial passages sent to the final re-ranker, and consequently the number of calls to LLM during the re-ranking phase, will decrease. This enhancement will improve the accuracy of the re-ranking step. Specifically, we have:

\begin{equation}
\sum_{i=1}^{N'} p_i \leq \sum_{i=1}^{N} p_i
\end{equation}

\begin{figure}[t]
    \centering
     \includegraphics[width=8cm]{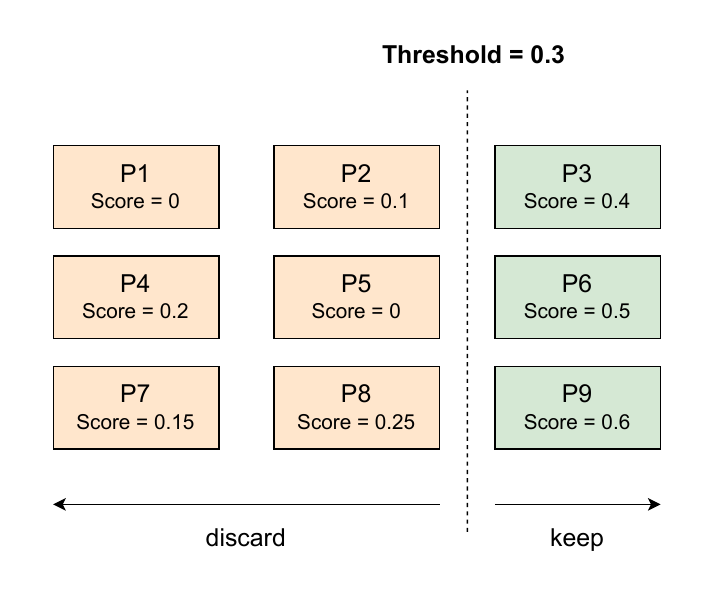}
    \caption{ The effect of the threshold on the number of passages.}
    \label{filtering}
\end{figure}

This demonstrates that the number of filtered passages, N', is either smaller or, at worst, equal to the initial number of retrieved passages N. Figure \ref{filtering} illustrates the effect of the threshold on retaining or discarding passages.

We test our proposed approach using an open-source LLM (i.e., Mixtral) which is easily accessible for both academic research and industry applications. Since this is an open-source model, there is no need for commercial LLM APIs, which can be expensive and may not satisfy some data privacy concerns. Furthermore, our experiments show that our approach allows smaller, more limited models, to remain competitive with much more demanding models. This can allow resource-constrained situations to still make use of state-of-the-art re-ranking. Our pre-filtering step is designed based on Zero-Shot prompting and thus eliminates the need to retrain or fine-tune the LLM. The complexity of our method is linear, O(n), and by discarding irrelevant passages, it reduces the number of LLM inferences required in the final re-ranking step. As access to the output logits of the model is not feasible with many LLMs, particularly closed-source ones, our method leverages the generation capabilities of the LLMs. Table~\ref{tab:Properties} presents properties of different re-ranking methods with LLMs.


\begin{table*}[ht]
\centering
\begin{tabular}{p{3.5cm} p{2.8cm} p{1.5cm} p{1.5cm} p{1.cm}}
\hline    
\textbf{Methods} & \#LLMcalls & Logits  & Batching & Generate \\
\hline
Pointwise & {\footnotesize $O(N)$} & {\footnotesize $\times$} & {\footnotesize $\times$} \\
Listwise & {\footnotesize $O(K * N)$} & {\footnotesize \ } & {\footnotesize \ }  & {\footnotesize $\times$}\\
Pairwise(all pairs) & {\footnotesize $O(N^2)$} & {\footnotesize $\times$} & {\footnotesize $\times$}  & {\footnotesize $\times$}\\
Pairwise(heap sort) & {\footnotesize $O(K * \log_2 N)$} & {\footnotesize $\times$} & {\footnotesize \ }  & {\footnotesize $\times$} \\
Pairwise(bubble sort) & {\footnotesize $O(K * N)$} & {\footnotesize $\times$} & {\footnotesize \ }  & {\footnotesize $\times$} \\
Setwise(heap sort) & {\footnotesize $O(K * \log_c N)$} & {\footnotesize $\times$} & {\footnotesize \ }  & {\footnotesize $\times$} \\
Setwise(bubble sort) & {\footnotesize $O\left( K * \left( \frac{N}{c - 1} \right) \right)$} & {\footnotesize $\times$} & {\footnotesize \ }  & {\footnotesize $\times$} \\
\textbf{Pre-Filtering} & {\footnotesize $O(N) + O(K * N')$} & {\footnotesize \ }   & {\footnotesize $\times$} & {\footnotesize $\times$} \\
\hline
\end{tabular}
\caption{Properties of different re-ranking methods with LLMs. \#LLM calls: the number of LLM API Calls in the worst case. Logits: access to the LLM's logits is required. Batching: batch inference is allowed. Generate: Token generation is required. N: the number of passages to re-rank. K: the number of top-k relevant passages to find. c: the number of compared passages at each step. N': the number of filtered passages, which is often much smaller than the initial N, and in the worst case, it is equal to N.}
\label{tab:Properties}
\end{table*}

\section{Passage Re-Ranking with LLMs}
After filtering out irrelevant passages, i.e., passages below the threshold, any LLM re-ranking method—Listwise, Pairwise, and Setwise—is applicable. For our experiments, we use Listwise prompting \cite{sun2023chatgpt}, which employs an instructional permutation generation method combined with a sliding window strategy to directly output a ranked list of candidate passages. In Listwise prompting, the LLMs receive a prompt with a given query, a list of candidate passages, and an instruction to generate a ranked list of passage labels based on their relevance to the query. 

Due to the input length limitations of LLMs, it is not possible to include all candidate passages in a single prompt. To handle this issue, this approach uses a sliding window method which involves ranking a window of candidate passages, starting from the bottom of the original ranking list and moving upwards. This process can be repeated multiple times to achieve an improved final ranking.
We select this approach as the final re-ranking step for two reasons; (1) this approach strikes a middle ground between efficiency and effectiveness, and (2) this method enables the use of early stopping mechanisms that focus specifically on the top-K rankings, enhancing efficiency.

\begin{table*}[!ht]
\centering
\vspace{-0.5em}
\begin{tabular}{lccccccccccc}
\hline
\textbf{Methods} & \textbf{Threshold} & \textbf{DL19} & \textbf{DL20} & \textbf{Covid} & \textbf{Touche} & \textbf{Signal} & \textbf{News} \\
\hline
BM25 & NA & 50.58 & 47.96 & 59.47 & 44.22 & 33.05 & 39.52 \\
\hline

\multicolumn{8}{c}{\textbf{Supervised}} \\
\hline

monoBERT (340M) & NA & 70.50 & 67.28 & 70.01 & 31.75 & 31.44 & 44.62 \\
monoT5 (220M) & NA & 71.48 & 66.99 & 78.34 & 30.82 & 31.67 & 46.83 \\
monoT5 (3B) & NA & 71.83 & 68.89 & 80.71 & 32.41 & 32.55 & 48.49 \\
RankT5 (3B) & NA & 72.95 & 69.63 & 82.00 & 37.62 & 31.80 & 48.15 \\
Cohere Rerank-v2 & NA & 73.22 & 67.08 & 81.81 & 32.51 & 29.60 & 47.59 \\
\hline

\multicolumn{8}{c}{\textbf{Unsupervised LLM-based}} \\
\hline

UPR (FLAN-T5-XXL) & NA & 62.00 & 60.34 & 72.64 & 21.56 & 30.81 & 42.99 \\
RG (FLAN-UL2) & NA & 64.61 & 65.39 & 70.22  & 24.67 & 29.68 &  43.78 \\
RankGPT (gpt-3.5-turbo) & NA & 65.80  & 62.91 & 76.67 & 36.18 & 32.12  & 48.85 \\
RankGPT (gpt-4) & NA & 75.59 & 70.56 & 85.51 & 38.57 & 34.40 & 52.89  \\
PRP-Allpair (FLAN-T5-XL) & NA & 69.75 & 68.12 & 81.86 & 26.93 & 32.08 & 46.52 \\
PRP-Sorting (FLAN-T5-XL) & NA & 69.28 & 65.87 & 80.41 & 28.23 & 30.95  & 42.95 \\
PRP-Allpair (FLAN-T5-XXL) & NA & 69.87 & 69.85 & 79.62 & 29.81 & 32.22  & 47.68 \\
PRP-Sliding-10 (FLAN-T5-XXL) & NA & 67.00 & 67.35 & 74.39 & 41.60 & 35.12 & 47.26 \\
PRP-Allpair (FLAN-UL2) & NA & 72.42  & 70.68  & 82.30  & 29.71 & 32.26 & 48.04 \\
PRP-Sorting (FLAN-UL2) & NA & 71.88 & 69.43 & 82.29 & 25.80 & 32.04 & 45.37 \\
setwise.heapsort (Flan-t5-large) & NA & 67.0 & 61.8 & 76.8 & 30.3 & 31.9 & 43.9 \\
setwise.bubblesort (Flan-t5-large) & NA & 67.8 & 62.4 & 76.1 & 39.4 & 35.1 & 44.7 \\
setwise.heapsort (Flan-t5-xl) & NA & 69.3 & 67.8 & 75.7 & 28.3 & 31.4 & 46.5 \\
setwise.bubblesort (Flan-t5-xl) & NA & 70.5 & 67.6 & 75.6 & 33.0 & 36.2 & 49.7 \\
setwise.heapsort (Flan-t5-xxl) & NA & 70.6 & 68.8 & 75.2 & 29.7 & 32.1 & 47.3 \\
setwise.bubblesort (Flan-t5-xxl) & NA & 71.1 & 68.6 & 76.8 & 38.8 & 34.3 & 47.9 \\
\hline

\multicolumn{8}{c}{\textbf{Pre-Filtering Step}} \\
\hline
PF (Mixtral-8x7B-Instruct) & 0.3 & 69.39 & 64.42 & 81.64 & \textbf{43.94} & \textbf{37.09} & 51.20  \\
PF (Mixtral-8x7B-Instruct) & 0.6 & 71.09 & - & - & - & - & - \\
PF (Mixtral-8x7B-Instruct) & 0.7 &   -   & 67.35 & - & - & - & - \\
\hline
\multicolumn{8}{c}{\textbf{Baseline Without Pre-Filtering Step}} \\
\hline
Mixtral-8x7B-Instruct & NA & 60.88 & 55.85 & 66.48 & 43.1 & 35.68  & 47.16  \\
\hline
\end{tabular}
\caption{Results (nDCG@10) on TREC and BEIR datasets by re-ranking top 100 documents retrieved by BM25.}
\label{tab:TREC}
\end{table*}
\vspace{-0.5em}

\section{Experimental Results of LLMs}

\subsection{Datasets}
Consistent with previous related research ~\cite{qin2023large, sun2023chatgpt, zhuang2023setwise}, our experiments are conducted on two well-established benchmark datasets in information retrieval research. These benchmark datasets include TREC-DL ~\cite{craswell2020overview,craswell2021overview2020} and BEIR ~\cite{thakur2021beir}.

\subsubsection{TREC}
TREC is a widely used benchmark dataset in information retrieval studies. To allow comparison with prior work, we use the test sets of the 2019 and 2020 competitions: TREC-DL2019 and TREC-DL2020. Both datasets are human-labeled and contain 43 and 54 queries, respectively. Each dataset is derived from the MS MARCO v1 passage corpus, which contains 8.8 million passages, with more comprehensive labeling. Based on the interpretation of the relevancy scores in the qrels files of these datasets, while passages with the relevancy level of \textbf{[{}1{}]} are considered irrelevant, they contribute positively to the nDCG score. Therefore, for these datasets, we conduct our experiment in two different scenarios with two different thresholds:
{(\romannumeral 1)} Considering passages with a relevancy level of \textbf{[{}1{}]} as relevant. 
{(\romannumeral 2)} Considering passages with a relevancy level of \textbf{[{}1{}]} as irrelevant.

\subsubsection{BEIR}
BEIR consists of diverse retrieval tasks and domains. Due to limited resources, we could not run our experiments on all of the BEIR tasks. Therefore, we choose to concentrate on four tasks in BEIR to evaluate the models:{(\romannumeral 1)} Covid retrieves scientific articles addressing queries related to COVID-19. {(\romannumeral 2)} Touche is a dataset that focuses on argument retrieval for controversial questions. {(\romannumeral 3)} Signal is a data collection of retrieved tweets for news articles.  {(\romannumeral 4)} News is a dataset that focuses on relevant news articles based on news headlines. In all of these four datasets, passages with the relevancy level of \textbf{[{}1{}]} are interpreted as relevant or partially relevant. Thus, we run our experiment only in one scenario with one threshold for these datasets: Considering passages with a relevancy level of \textbf{[{}1{}]} as relevant.

\subsection{Implementation and Metrics}
To enable fair and direct comparison with prior works, our experiments are conducted using the top 100 passages retrieved by BM25, serving as the first-stage retriever through Pyserini\footnote{https://github.com/castorini/pyserini} with its default settings. We evaluate the effectiveness of approaches using the NDCG@10 metric, which is the official evaluation metric for the datasets used.
As one of the main goals of this paper is to investigate the effects of open-source LLMs on re-ranking tasks, we employ the open-source language model Mixtral-8x7B-Instruct-v0.1~\cite{jiang2024mixtral}, which has 46.7 billion parameters, for both the LLM-based pre-filtering and re-ranking steps.
Due to the input length limitations of LLMs, we divide the initial list of passages into smaller chunks for the pre-filtering step, each containing 5 elements, before processing them with the LLM. We use 5 elements as it provides a balance between computational efficiency and effectiveness. For the re-ranking step,  we implement the sliding window strategy introduced by Sun et al.~\cite{sun2023chatgpt}, with a window size of 10 and a step size of 5.

We carry out our experiment on a Google Cloud a2-highgpu-1g machine equipped with a single NVIDIA A100 40GB GPU with 40 GB of memory, and 12 vCPUs. Due to resource constraints, the LLM is loaded with 4-bit quantization. Using these resources, we conducted our experiments separately for each group of datasets using the methodology presented in Section \ref{sec:Methodology}. 

\subsection{Setting the Threshold Values}

As discussed in Section~\ref{sec:threshold}, our approach depends on threshold values.
For all four tasks in the BEIR benchmark, we set a single threshold by considering passages with a relevancy level of \textbf{[{}1{}]} as relevant. Our analysis determines that \textbf{0.3} is the optimal threshold for BEIR, as it yields the highest F1 score compared to other values.

For both datasets in the TREC benchmark, we set two thresholds: one by considering passages with a relevancy level of \textbf{[{}1{}]} as relevant, and another by considering passages with a relevancy level of \textbf{[{}1{}]} as irrelevant. Here, we find that thresholds of \textbf{0.6} and \textbf{0.7} are respectively optimal, achieving the highest F1 scores compared to other values.

Our evaluation of these threshold values answers our first research question (\textbf{RQ1}). We find that even a small percentage of qrels (8\% of the total dataset) is enough to determine a threshold value that can effectively help our chosen LLM filter out irrelevant passages. This implies that limited effort from expert annotators is necessary to allow our approach to work (and improve the state of the art). Furthermore, we find that the threshold 0.3 (or its inverse of 0.7) is stable for our LLM across most datasets. This implies that future datasets need not necessarily have expert judgement (or qrels) to use our approach in a useful fashion. The use of a previously determined threshold may be sufficient to obtain decent pre-filtering power. 

\subsection{Results on Benchmarks}
To situate our results, we compare our approach with state-of-the-art supervised and unsupervised passage re-ranking techniques. The supervised baselines are as follows: \textbf{monoBERT}~\cite{nogueira2019multi}: A BERT-large based cross-encoder re-ranker, trained using the MS MARCO dataset.
\textbf{monoT5}~\cite{nogueira-etal-2020-document}: A sequence-to-sequence re-ranker that uses T5 to calculate the relevance scores using pointwise ranking loss.
\textbf{RankT5}~\cite{zhuang2023rankt5}: A re-ranker that employs T5 and uses listwise ranking loss.
\textbf{Cohere Rerank}\footnote{https://txt.cohere.com/rerank/}: A passage reranking API named rerank-english-v2.0, developed by Cohere\footnote{https://cohere.com/rerank}, which does not explain the architecture or training method of the model. The unsupervised LLM-based baselines include:
\textbf{Unsupervied Passage Re-ranker (UPR)}~\cite{sachan2022improving}: The pointwise approach with instructional query generation. \textbf{Relevance Generation (RG)}~\cite{liang2022holistic}: The pointwise approach that generates relevance judgments for a given query and candidate items.
\textbf{RankGPT}~\cite{sun2023chatgpt}: The listwise approach generates a ranked list of passage labels based on their relevance to the query. \textbf{Pairwise Ranking Prompting (PRP)}~\cite{qin2023large}: The pairwise approach involves generating the label for the passage that is more relevant to the query. \textbf{Setwise Approach}~\cite{zhuang2023setwise}: The setwise approach, which accelerates the sorting algorithms used in the pairwise method. We also compare our results to \textbf{Mixtral-8x7B-Instruct} without our pre-filtering step to show the improvement obtained through our pre-filtering.

Table ~\ref{tab:TREC} presents the evaluation results obtained from the TREC and BEIR datasets. Results show that: {(\romannumeral 1)}: The pre-filtering method can achieve the best results on the Signal and Touche datasets for NDCG@10, even outperforming commercial solutions (e.g., GPT-4). {(\romannumeral 2)}: The pre-filtering method outperforms all other methods, other than BM25, for the Touche dataset.{(\romannumeral 3)}: Pre-filtering achieves an average improvement of 1.64 in nDCG@10 on TREC compared to unsupervised LLM-based methods. {(\romannumeral 4)}: Pre-filtering achieves an average improvement of 7.2\% in nDCG@10 on all datasets compared to our baseline without pre-filtering. These results answer our \textbf{RQ2}, and show that pre-filtering irrelevant passages before re-ranking improves its overall results.

\section{conclusion}
In this paper, we conduct a study on the use of a pre-filtering step before passage re-ranking with LLMs. We introduce a novel approach to further exploit the power of LLMs in IR passage ranking. Our experiments on three benchmarks (TREC-DL2019, TREC-DL2020, and four BEIR tasks) show that using our approach, smaller LLMs (i.e., Mixtral-8x7B-Instruct with 4-bit quantization), can be made competitive with much larger models. While our approach does require some expert input, we also show that the amount of input needed is small, and furthermore, that the thresholds used appear to be effective across multiple benchmarks.

\newpage
\section*{Limitations}

The limitations of this work include the threshold-setting process, where the initial threshold value is selected randomly. Based on this value, the F1 score is computed to find the optimized threshold. Although the thresholds introduced in this paper are derived from analysis and the results reveal their effectiveness, these thresholds are dependent on the dataset and might differ for different datasets. Additionally, our approach depends heavily on the qrels files and the interpretation of relevance levels for the dataset. As the pre-filtering step is an intermediate step in the information retrieval pipeline, its effectiveness is closely related to other elements in the pipeline, such as the retriever and re-ranker. Due to hardware constraints, particularly GPUs, we only ran our experiments on four datasets (Covid, News, Signal, Touche) instead of the eight datasets used in BEIR, which are referenced in similar works. We also tested our experiments on other open-source models, such as Llama-2-13b, but while the approach did provide improvements over non-prefiltered results, the results were not comparable to the current state-of-the-art.

\section*{Ethics Statement}

We acknowledge the importance of the ACM Code of Ethics and fully agree with its principles. All our results are derived from our own code, which we have made publicly accessible. We fully acknowledge the weaknesses, risks, and potential harms associated LLMs, which can lead to various issues during their use. Throughout the development of this paper, we encountered several challenges with LLMs, such as incorrect result generation and occasional failures to produce answers. Furthermore, there are instances where LLMs exhibit bias towards certain passages, indicating that their use in critical tasks can be problematic. Such biases and the generation of incorrect information, known as hallucinations, underscore the importance of caution when employing LLMs. We have cited all methods and techniques used from other papers and research. In our work with Mixtral and LLaMA2, we adhered to Hugging Face's policies and conditions for Mixtral and complied with Meta's licensing requirements for LLaMA2.

\bibliography{emnlp2023}
\bibliographystyle{acl_natbib}

\appendix



\end{document}